\documentclass{article}

     \usepackage[final,nonatbib]{neurips_2020}

\usepackage{wrapfig}

\usepackage[utf8]{inputenc} 
\usepackage[T1]{fontenc}    
\usepackage{hyperref}       
\usepackage{url}            
\usepackage{booktabs}       
\usepackage{amsfonts}       
\usepackage{nicefrac}       
\usepackage{microtype}      
\usepackage{mathtools}
\usepackage{mathtools}
\usepackage{dsfont}
\usepackage[dvipsnames]{xcolor}
\usepackage[colorinlistoftodos]{todonotes}
\usepackage{booktabs}
\usepackage{xfrac}
\usepackage{bbm}
\usepackage{algorithm}
\usepackage{algpseudocode}
\usepackage[most]{tcolorbox}
\usepackage{xparse}
\usepackage{lipsum}
\usepackage{changepage}
\usepackage{enumitem}

\usepackage{sidecap}

\newcommand{\squishlist}{
   \begin{list}{$\bullet$}
    { \setlength{\itemsep}{0pt}      \setlength{\parsep}{3pt}
      \setlength{\topsep}{3pt}       \setlength{\partopsep}{0pt}
      \setlength{\leftmargin}{1.5em} \setlength{\labelwidth}{1em}
      \setlength{\labelsep}{0.5em} } }
\newcommand{\squishlisttwo}{
   \begin{list}{$\bullet$}
    { \setlength{\itemsep}{0pt}    \setlength{\parsep}{0pt}
      \setlength{\topsep}{0pt}     \setlength{\partopsep}{0pt}
      \setlength{\leftmargin}{2em} \setlength{\labelwidth}{1.5em}
      \setlength{\labelsep}{0.5em} } }
\newcommand{\squishend}{
    \end{list}  }

\DeclareMathOperator*{\argmin}{arg\,min}

\newcommand{\gauss}{{\cal N}}

\newcommand{\expect}[2]{\mathds{E}_{{#1}} \left[ {#2} \right]}

\newcommand{\diag}{\mbox{diag}}

\newcommand{\cC}{{\cal C}}

\DeclareMathAlphabet{\mathpzc}{OT1}{pzc}{m}{n}

\newcommand{\norm}[1]{\left\lVert#1\right\rVert_1}
\newcommand{\normtwo}[1]{\left\lVert#1\right\rVert_2}
\newcommand{\wsum}{\norm{\omega}}

\title{Gaussian Gated Linear Networks}

\author{
  David Budden\thanks{$^\dag$ Equal contributions.} \quad Adam H. Marblestone$^*$ \quad  Eren Sezener$^*$\\
  \textbf{Tor Lattimore} \quad \textbf{Greg Wayne}$^\dag$ \quad \textbf{Joel Veness}$^\dag$ \vspace{2mm}\\
  DeepMind\vspace{1mm}\\
  \texttt{aixi@google.com}\\
}

\begin{document}

\maketitle
\vspace{-3mm}
\begin{abstract}
We propose the Gaussian Gated Linear Network (G-GLN), an extension to the recently proposed GLN family of deep neural networks.
Instead of using backpropagation to learn features, GLNs have a distributed and local credit assignment mechanism based on optimizing a convex objective. 
This gives rise to many desirable properties including universality, data-efficient online learning, trivial interpretability and robustness to catastrophic forgetting. 
We extend the GLN framework from classification to multiple regression and density modelling by generalizing geometric mixing to a product of Gaussian densities. The G-GLN achieves competitive or state-of-the-art performance on several univariate and multivariate regression benchmarks, and we demonstrate its applicability to practical tasks including online contextual bandits and density estimation via denoising. 
\end{abstract}

\vspace{-3mm}
\section{Introduction}
\label{sec:intro}
Recent studies have demonstrated that backpropagation-free deep learning, particularly the Gated Linear Network (GLN) family~\cite{sezener2020online,veness2017online,veness2019gated}, can yield surprisingly powerful models for solving classification tasks. This is particularly true in the online regime where data efficiency is paramount. In this paper we extend GLNs to model real-valued and multi-dimensional data, and demonstrate that their theoretical and empirical advantages apply to far broader domains than previously anticipated.

The distinguishing feature of a GLN is distributed and local credit assignment. A GLN associates a separate convex loss to each neuron such that all neurons (1) predict the target distribution directly, and (2) are optimized locally using online gradient descent. A half-space ``context function'' is applied per neuron to select which weights to apply as a function of the input features, allowing the GLN to learn highly nonlinear functions. This architecture gives rise to many desirable properties previously shown in a classification setting: (1) trivial interpretability given its piecewise linear structure, (2) exceptional robustness to catastrophic forgetting, and (3) provably universal learning; a sufficiently large GLN can model any well-behaved, compactly supported density function to any accuracy, and any no-regret convex optimization method will converge to the correct solution given enough data.

\paragraph{Related Work.}
We extend the previous Bernoulli GLN (B-GLN) formulation to model multivariate, real-valued data by reformulating the GLN neuron as a gated product of Gaussians. This Gaussian Gated Linear Network (G-GLN) formulation exploits the fact that exponential family densities are closed under multiplication~\cite{welling2005exponential}, a property that has seen much use in Gaussian Process and related literature~\cite{williams2002products, peng2019mcp, cao2014generalized, marblestone2020product, deisenroth2015distributed}. Similar to the B-GLN, every neuron in our G-GLN directly predicts the target distribution. This idea is shared with work in supervised learning where targets are predicted from intermediate layers. The motivations for local, layer-specific training include improving gradient propagation and representation learning \cite{lee2015deeply, rasmus2015semi, lowe2019putting,lee2018gradient}, decoding for representation analysis \cite{alain2016understanding} and making neural networks more biologically plausible \cite{sussillo2009generating, nokland2019training, mostafa2018deep} by avoiding backpropagation. The use of context-dependent weight selection (gating) in the GLN algorithm family resembles proposals to improve the continual and multi-task learning properties of deep networks \cite{schmidhuber1992learning, ha2016hypernetworks, cheung2019superposition, von2019continual, perez2018film} by using a conditioning network to gate a principal network solving the task.

\paragraph{Paper Outline.}
We begin by reviewing some background on weighted products of Gaussian densities, and describe how the relevant weights can be adapted using well-known online convex programming techniques~\cite{Hazan16}.
We next show how to augment this adaptive form with a gating mechanism, inspired by earlier work on classification with GLNs \cite{veness2017online,veness2019gated}, which gives rise to the notion of neuron in G-GLNs.
We then introduce G-GLNs, feed-forward networks of locally trained neurons, each computing a weighted product of Gaussians with input-dependent, gated weights. We conclude by providing a comprehensive set of experimental results demonstrating the impressive performance of the G-GLN algorithm across a diverse set of regression benchmarks and practical applications including contextual bandits and image denoising.

\section{Background}
\label{sec:background}

The Gaussian distribution has a number of well-known properties that make it well suited for machine learning applications.
Here we briefly review two of these important properties: closure under multiplication and convexity with respect to its parameters under the logarithmic loss, which we will later exploit to define our notion of a G-GLN neuron. 

\subsection{Weighted Products of Gaussian Densities}
\label{subsec:pog}
A weighted product of Gaussians is closed in the sense that it yields another Gaussian. More formally, let $\mathbb{R}_+$ denote the set of non-negative real numbers. For notational simplicity, we first construct the univariate case. Let $\mathcal{N}(\mu, \sigma^2)$ denote the univariate Gaussian PDF with mean $\mu \in \mathbb{R}$ and standard deviation $\sigma \in \mathbb{R}_+$.
Now, given $m$ univariate Gaussian experts of the form $\mathcal{N}(\mu_1, \sigma^2_1)$, \dots, $\mathcal{N}(\mu_m, \sigma^2_m)$ with associated PDFs 
\begin{equation}
f_i(y) = \frac{1}{\sigma_i \sqrt{2 \pi} } \exp \left \{- \frac{1}{2} \left(\frac{y - \mu_i}{\sigma_i} \right)^2 \right \},
\end{equation}
and an $m$-dimensional vector of weights $w = (w_1, w_2, \dots, w_m) \subset \mathbb{R}_+^m$, we define a weighted Product of Gaussians (PoG) 
as
\begin{equation}\label{eq:pog}
\text{\sc PoG}_w(y \, ; \, f_1(\cdot), \dots, f_m(\cdot)) := \frac{1}{Z} \prod\limits^m_{i=1} [f_i(y)]^{w_i} 
\text{~~~~with~~~~}  Z:=\int \prod\limits^m_{i=1} [f_i(y)]^{w_i} \, dy.
\end{equation}

It is straightforward to show that this formulation gives rise to a Gaussian distribution whose mean and variance jointly depend on $w$; see Appendix \ref{app:pog_tutorial} for a short derivation.
In particular we can exactly interpret the weighted product of experts as another Gaussian expert $\mathcal{N}\left(\mu_\textsc{PoG}(w), \sigma^2_\textsc{PoG}(w) \right)$
where
\begin{equation}\label{eq:gaussin_magic2}
\sigma^2_\textsc{PoG}(w) := \left[ \sum_{i=1}^m  \frac{w_i}{\sigma^2_i} \right]^{-1}
\text{~~~~and~~~~}
\mu_\textsc{PoG}(w) := \sigma^2_\textsc{PoG}(w) \left[ \sum_{i=1}^m \frac{w_i \, \mu_i}{\sigma^2_i} \right].
\end{equation}

The same closure property holds for the multivariate case (e.g. see \cite{bromiley2018}).
Let $\mathcal{N}(\mu, \Sigma)$ denote the $d$-dimensional multivariate Gaussian PDF, with mean $\mu \in \mathbb{R}^d$ and covariance matrix $\Sigma \in \mathbb{R}^{d \times d}$, and let $\mathcal{I}_d$ denote the $d$-dimensional identity matrix.
In the general case, given $m$ multivariate $d$-dimensional Gaussian experts, $\mathcal{N}(\mu_1, \Sigma_1)$, \dots, $\mathcal{N}(\mu_m, \Sigma_m)$, we have
\begin{equation}\label{eq:gaussin_magic3}
\Sigma_\textsc{PoG}^{-1}(w) = \sum\limits_{i=1}^m w_i \Sigma_{i}^{-1} 
\text{~~~and~~~} 
\mu_\textsc{PoG}(w) = \Sigma_\textsc{PoG}(w) \sum\limits_{i=1}^m w_i \Sigma_{i}^{-1} \mu_i.
\end{equation}
Note that $\mu_\textsc{PoG}(w)$ is a convex combination of the means $\mu_i$ of its inputs, which implies that $\mu_\textsc{PoG}(w)$ must lie within the convex hull formed from all the $\mu_i$.
In the isotropic case with $\Sigma^{-1}_i = \tau_i \mathcal{I}_d$ for precision $\tau_i > 0$, Equation \ref{eq:gaussin_magic3}  simplifies to 
\begin{equation}\label{eq:gaussin_magic4}
\Sigma_\textsc{PoG}^{-1}(w) = \left( \sum\limits_{i=1}^m w_i \tau_i\right) \mathcal{I}_d 
\text{~~~and~~~} 
\mu_\textsc{PoG}(w) =  \left(\sum\limits_{i=1}^m w_i \tau_i\right)^{-1}  \sum_{i=1}^m w_i \tau_i  \mu_i.
\end{equation}
Note that if all the initial experts are isotropic, the product of Gaussians must also be isotropic.
Although less general, the isotropic form has considerable computational advantages for high-dimensional multivariate regression 
(since the inverses can be computed in $\mathcal{O}(d)$ time),
and will be used in our larger scale multivariate regression experiments.

\subsection{Online Convex Programming Formulation}
\label{subsec:ocp}
We now show how to adapt the weights in Equation \ref{eq:pog} using online convex programming.
Assuming a standard online learning setup under the logarithmic loss, we define the instantaneous loss given a target $y \in \mathbb{R}$ with respect to a fixed weight vector $w \in \mathbb{R}_+^m$ as
\begin{align}\label{eq:loss}
\ell(y ; w) & := -\log \text{\sc PoG}_w(y \,;\, f_1(y), \dots, f_m(y)) 
~\equiv~ \log \sigma_\textsc{PoG}^2(w) + \frac{ \left( y - \mu_\textsc{PoG}(w) \right)^2}{\sigma^2_\textsc{PoG}(w)},
\end{align}
with equivalence following by dropping non-essential constant terms.
It is straightforward to show $\ell(y ; w)$ is convex in $w$, either directly (as in Appendix \ref{app:properties_of_loss}), or by appealing to known properties of the log-partition function for exponential family members \cite{wainwright2008graphical}.

As we are interested in large scale applications, we derive an Online Gradient Descent (OGD) \cite{Zinkevich03} learning scheme to exploit the convexity of the loss in a principled fashion.
To apply OGD in our setting, we need to restrict the weights to a choice of compact convex set $\mathcal{W} \subset \mathbb{R}_+^m$.
For simplicity of exposition, we focus our presentation on the case where the weight space is defined as
\begin{equation}
\mathcal{W} := \{ w \in [0,b]^m \,:\, \norm w \geq \epsilon \},    
\end{equation}
where $0 < \epsilon < 1$ and $b \geq 1$.
As $\mathcal{W}$ is formed from the intersection of a scaled hypercube and a half-space, it is a convex set with finite diameter, and is clearly compact and non-empty.
OGD works by performing two operations, a gradient step and a projection of the modified weights back into $\mathcal{W}$ if the gradient update pushed them outside of $\mathcal{W}$.
This projection is essential, as it is responsible for both ensuring that the weighted product of Gaussians is well-defined (e.g. positive variance) and for providing no-regret guarantees comparable to what was previously achieved for B-GLNs~\cite{veness2017online}.

\section{G-GLN Neurons}
\label{sec:neuron}
We now introduce a new type of neuron which will constitute the basic learning primitive within a G-GLN.
The key idea is that further representational power can be added to a weighted product of Gaussians via a contextual gating procedure.
We achieve this by extending the previous weighted product of Gaussians model with an additional type of input, which we call \emph{side information}.
The side information will be used by a neuron to select a weight vector to apply for a given example from a table of weight vectors. 
In typical applications to regression, the side information is defined as the (normalized) input features for an input example: i.e. $z = (x-\bar{x})/\sigma_x$. 

More formally, associated with each neuron is a context function $c : \mathcal{Z} \to \cC$, where $\mathcal{Z}$ is the set of possible \textit{side information} and $\cC = \{0,\dots,k-1 \}$ for some $k \in \mathbb{N}$ is the \textit{context space}.
Each neuron $i$ is now parameterized by a weight matrix
$
W_i = \begin{bmatrix}
w_{i,0} \dots w_{i,k-1} 
\end{bmatrix}^\top
$
with each row vector $w_{ij} \in \mathcal{W}$ for $0 \leq j < k$.
The context function $c$ is responsible for mapping side information $z \in \mathcal{Z}$ to a particular row $w_{i,c(z)}$ of $W_i$, which we then use to weight the Product of Gaussians. 

In other words, a G-GLN neuron can be defined in terms of Equation \ref{eq:pog} by
\begin{equation}
\label{eq:gated_peg_defn}
\text{\sc PoG}^c_{W}(y \, ; \, f_1(\cdot), \dots, f_m(\cdot), z) := \text{\sc PoG}_{w^{c(z)}}(y \,;\, f_1(\cdot), \dots, f_m(\cdot)),
\end{equation}
with the associated loss function $-\log(\text{\sc PoG}^c_{W}(y \, ; \, f_1(y), \dots, f_m(y), z))$ inheriting all the properties needed to apply Online Convex Programming directly from Equation \ref{eq:loss}.

\paragraph{Half-space Gating.}
\label{sec:hyperplanedetails}
We restrict our attention to the class of half-space context functions, as in \cite{veness2017online}.
Given a normal vector $v \in \mathbb{R}^d$ and offset $b \in \mathbb{R}$, consider the associated affine hyperplane $\{z \in \mathbb{R}^d : z \cdot v = b \}$.
This divides $\mathbb{R}^d$ in two, giving rise to two half-spaces, one of which we denote $H_{v,b} = \{z \in \mathbb{R}^d : z \cdot v \geq b\}.$
The associated half-space context function is then given by $c(z) := 1$ if $z \in H_{v,b}$ or 0 otherwise.
Richer notions of context can be created by composition.
In particular, any finite set of $s$ context functions $\{ c_i : \mathcal{Z} \to \mathcal{C}_i \}_{i=1}^s$ with associated context spaces $\cC_1, \dots, \cC_s$ can be composed into a single higher order context function by defining $c(z) = (c_1(z),...,c_s(z))$.
We will refer to the choice of $s$ as the \emph{context dimension}. 

\paragraph{Bias Models.}
G-GLN neurons transform an input set of Gaussians to an output Gaussian.
Recall that the mean of a product of Gaussian PDFs must lie within the convex hull defined by the means of the individual input Gaussian PDFs (Section~\ref{subsec:pog}). To ensure the G-GLN neuron can represent any mean in $[-r,r]^D$, where $D$ is the target dimension, we therefore concatenate a number of \emph{bias inputs}, i.e. constant Gaussian PDFs to the input of each neuron. In the univariate case, we concatenate two Gaussian PDFs with mean $\pm r$ with a typical value of $r=5$ (the target is standardized). This generalizes to the multivariate case by multiplying the two scalars $\pm rD$ against each $D$-dimensional standard basis vector, allowing the convex hull of the bias inputs to span the $[-r,r]^D$ target hypercube.

\section{G-GLN Architecture}
\label{sec:ggln}
\begin{figure}[t]
\centering
\includegraphics[width=\linewidth]{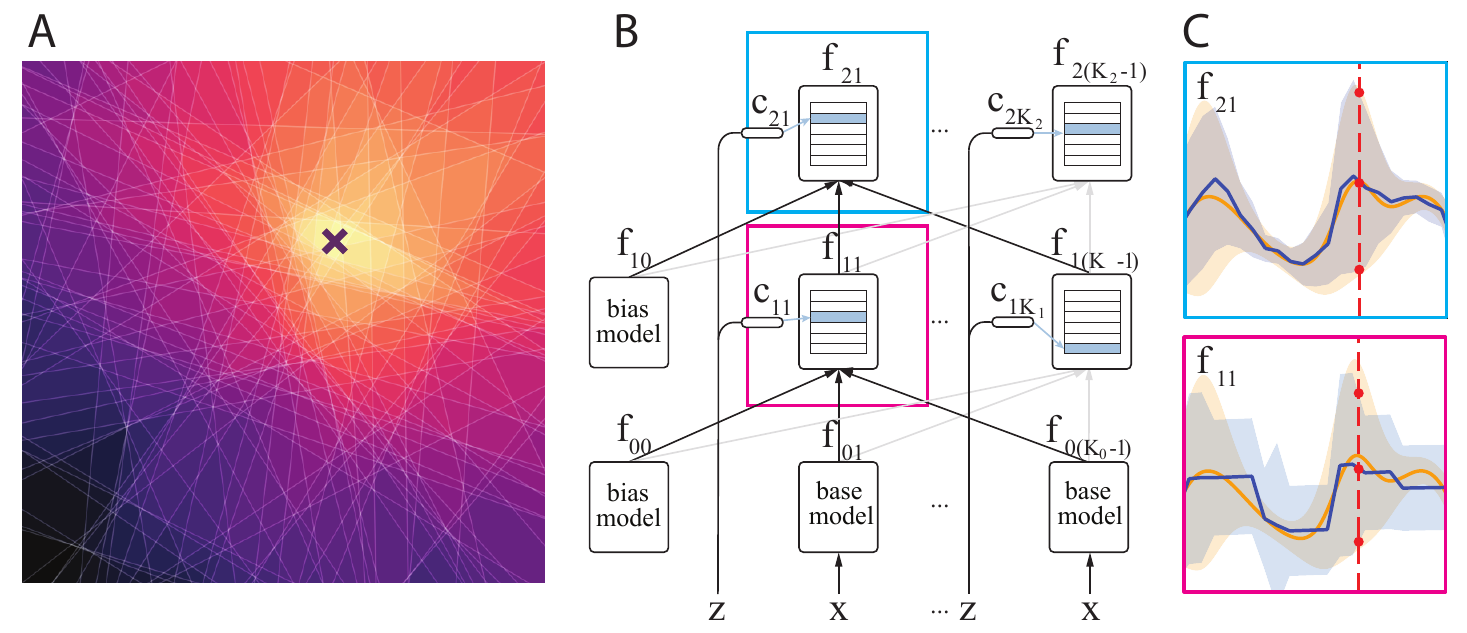}
\caption{
(\textbf{A}) Illustration of half-space gating for a 2D context. Color represents how many half-spaces intersect with the data point $x$. Within each region of constant color (each polytope), the gated weights for a G-GLN network are constant. (\textbf{B}) G-GLN feed-forward architecture. Each neuron uses its active weights to predict the target density as a function of the preceding layer outputs. (\textbf{C}) Illustration of the function sufficient statistics (mean and standard deviation) predicted by two neurons at different G-GLN layers, visualized for both for a single input (red line) and across all inputs within a fixed range (blue). Deeper neurons more accurately reconstruct the true density (orange).
}
\label{fig:ggln}
\end{figure}

We now describe how the neurons defined in the previous section are assembled to form a G-GLN (Figure 1, B). Similar to its B-GLN predecessor~\cite{veness2017online,veness2019gated}, a G-GLN is a feed-forward network of data-dependent distributions. 
Each neuron calculates the sufficient statistics ($\mu, \sigma^2$) for its associated PDF using its active weights, given those emitted by neurons in the preceding layer. 

\paragraph{Inputs and Side Information.}
There are two types of input to neurons in the network. The first is the side information, which can be thought of as the input features, and is used to determine the weights used by each neuron via half-space gating. The second is the input to the neuron, which will be the PDFs output by the previous layer, or in the case of layer 0, some provided base models.
To apply a G-GLN in a supervised learning setting, we need to map the sequence of input-label pairs $(x_t, y_t)$ for $t=1,2,\dots$ onto a sequence of (side information, base Gaussian PDFs, label) triplets $(z_t, \{ f_{0i} \}_{i}, y_t)$.  
The side information $z_t$ will be set to the (potentially normalized) input features $x_t$.
The Gaussian PDFs for layer 0 will generally include the necessary base Gaussian PDFs to span the target range, and optionally some base prediction PDFs that capture domain-specific knowledge. 

\paragraph{Model Description.}
More formally, a G-GLN consists of $L+1$ layers indexed by $i \in \{0,\ldots,L\}$, with $K_i$ neurons in each layer. The weight space for a neuron in layer $i$ will be denoted by $\mathcal{W}_i$; the subscript is needed since the dimension of the weight space depends on $K_{i-1}$.
Each neuron/distribution will be indexed by its position in the network when laid out on a grid; for example, $f_{ik}$ will refer to the family of PDFs defined by the $k$th neuron in the $i$th layer.
Similarly, $c_{ik}$ will refer to the context function associated with each neuron in layers $i \geq 1$, and $\mu_{ik}$ and $\sigma^2_{ik}$ (or $\Sigma_{ik}$ in the multivariate case) referring to the sufficient statistics for each Gaussian PDF.

\paragraph{Heteroskedastic Regression Example.}
We show an illustrative example on a popular heteroskedastic benchmark function $\mathcal{N}(\mu(x_i), \exp(g(x)))$, with mean $\mu(x) = 2 [\exp(-30(x - 0.25)^2) + \sin(\pi x^2)] - 2$ and the logarithm of the standard deviation $g(x) = \sin(2\pi x)$~\cite{silverman85,kersting2007most}. 
Intermediate layer outputs in the G-GLN are illustrated in Figure~\ref{fig:ggln}(C).
For each training input $x$ (red line), with target $y$ (intersection of dashed red line and yellow curve), and for each neuron: (1) a set of active weights are selected by applying the context function to the broadcast side information (in this case simply $x$), (2) the active weights are used to predict the target distribution as a function of preceding predictions, and (3) the active weights are updated with respect to the loss function defined in Equation (\ref{eq:loss}). Figure~\ref{fig:ggln}(B) compares the predictions (blue) for all values of $x$ for two individual neurons. It is clearly evident from inspection that neurons in higher layers produce more accurate predictions of the sufficient statistics given only the preceding predictions as input.

\paragraph{Generating Context Functions.}
We sample our context functions randomly according to the scheme first introduced in \cite{veness2017online,veness2019gated}, which is inspired by the SimHash method \cite{Charikar2002} for locality sensitive hashing.
Recall that a half-space context is defined by $H_{v,b}$; to sample $v$, we first generate an i.i.d. random vector $x = (x_1, ..., x_d)$ of dimension $d$, with each component of $x$ distributed according to the unit normal $\mathcal{N}(0,1)$, and then divide by its $2$-norm, giving us a vector $v = x / ||x||_2$. This scheme uniformly samples points from the surface of a unit sphere. 
The scalar $b$ is sampled directly from a standard normal distribution.

To gain intuition for this procedure, consider Figure~\ref{fig:ggln}(A).
There is a 1-1 mapping between any convex polytope formed from the intersection of each of the half-spaces, and the collective firing pattern of all context functions in the network.
Choices of side information close in terms of cosine similarity will map to similar sets of weights.
A (local) update of the weights corresponding to a particular convex region will therefore affect neighbouring regions, but with decreasing impact in proportion to the number of overlapping half-spaces. 

\section{G-GLN Algorithm}
\begin{algorithm}[t!]
\begin{algorithmic}[1]
\caption{G-GLN: inference with optional update}\label{alg:ggln}
\medskip
\item {\bfseries Input:} base model / features $\{ \mu_{0j}, \sigma_{0j}\}_{j=0}^{K_0-1}$
\item {\bfseries Input:} side information $z \in \mathcal{Z}$, target $y \in \mathbb{R}$
\item {\bfseries Input:} G-GLN weights $\{W_{ik}\}$, learning rate $\eta \in (0,1)$
\medskip
\item {\bfseries Output:} Gaussian PDF 
\medskip
\For{$i \in \{ 1,\dots,L \}$}
\For{$k \in \{ 1, \dots, K_i \}$}
\State $(w_0, \dots, w_{K_{i-1}}) \leftarrow W_{ik c_{ik}}(z)$
\State $\sigma^2_{ik} \leftarrow \left[ \sum_{j=0}^{K_{i-1}} w_j/\sigma_{i-1,j}^2 \right]^{-1}$\label{alg:sigma_update}
\State $\mu_{ik} \leftarrow \sigma^2_{ik} \left[ \sum_{j=0}^{K_{i-1}} w_j \mu_{i-1,j}/\sigma_{i-1,j}^2 \right]$\label{alg:mu_update}
    \State $W_{ik c_{ik}(z)} \leftarrow \textsc{Proj}_i[W_{ik c_{ik}(z)} - \eta \nabla \ell_{ik}(y ; z)]$ // \text{(if learning)}
\EndFor
\EndFor
\medskip
\State \Return  $\gauss(\mu_{L1}$, $\sigma^2_{L1})$
\end{algorithmic}
\end{algorithm}

We now describe how inference is performed in a G-GLN. For layer 0, we assume all the base models are given. For layers $i \geq 1$, we then have
\begin{equation}\label{eq:ggln_rec}
f_{ik}(y \,;\, z) :=  \text{\sc PoG}^{c_{ik}}_{W_{ik}}\left(y \, ; \, f_{i-1, 0}(\cdot \,;\, z), \dots, f_{i-1, K-1}(\cdot \,;\, z), z \right).    
\end{equation}

Equation \ref{eq:ggln_rec} makes it explicit that, conceptually, a G-GLN is a network of Gaussian PDFs, each of which depend on the side information $z$ via gating.
Computationally, this involves a forward pass of the network to compute the relevant sufficient statistics for each neuron (using Equations \ref{eq:gaussin_magic2}-\ref{eq:gaussin_magic4}).
By re-expressing Equation \ref{eq:ggln_rec} as 
\begin{equation*}
f_{ik}(y \,;\, z) \propto \exp \left\{ \log \prod_{j=1}^{K_{i-1}} [f_{i-1,j}(y \,;\, z)]^{W_{ij c_{ik}(z)}}  \right\} 
= \exp \left\{ \sum_{j=1}^{K_{i-1}} W_{ij c_{ik}(z)} \log \left ( f_{i-1,j}\left(y \,;\, z\right) \right) \right\}, 
\end{equation*}
one can view each neuron as having an exponential output non-linearity and a logarithic input non-linearity.
Since these non-linearities are inverses of each other, stacking layers causes the non-linearities to cancel, so the density output by a G-GLN collapses to a linear function of the gated weights (i.e. a Gated Linear Network).
The same cancellation argument applies to B-GLN~\cite{veness2017online}, where the output and input non-linearities are the sigmoid and logit functions. 

A distinguishing feature of a G-GLN is that every neuron directly attempts to predict the target, by locally boosting the accuracy of its input distributions.
Because of this, every neuron will have its own loss function defined only in terms of its own weights.
Given a (potentially vector-valued) target $y$, and side information $z$ (which will typically be identified with the input features), each neuron-specific loss function will be
\begin{equation}
\ell_{ik}(y ; z) := -\log f_{ik} \left(y \,;\, z \right) .
\end{equation}
This loss can be optimized using online gradient descent \cite{Zinkevich03}, which involves performing a step of gradient descent, and projecting the weights back onto $\mathcal{W}_i$, via the update rule
\begin{equation}
W_{ik c_{ik}(z)} \leftarrow \textsc{Proj}_i[W_{ik c_{ik}(z)} - \eta \nabla \ell_{ik}(y ; z)],    
\end{equation}
where $W_{ikj}$ refers to the $j$th row of the neurons weight matrix $W_{ik}$, $\eta > 0$ is the learning rate and $\textsc{Proj}_i[w] := \argmin_{w' \in \mathcal{W}_i } \normtwo{w' - w}$ is the projection operator with respect to the Euclidean norm.

Algorithm~\ref{alg:ggln} provides pseuodocode for both inference and (optionally) weight adaptation for a univariate G-GLN for a given input, with the top-most neuron taken as the final Gaussian PDF.
The multivariate case can be obtained by replacing lines 8-9 with Equation~\ref{eq:gaussin_magic3} or \ref{eq:gaussin_magic4}.
The total time complexity to perform inference is the sum of the cost of computing the gating operations $O \left( d \left( \sum_{i=1}^L K_i \right) \right) $, where $d$ is the dimensionality of the input vector, and the cost of propagating the sufficient statistics through the network, $O \left( \sum_{i=1}^L K_i K_{i-1} \right)$.

\section{Experimental Results}
\label{sec:results}
\begingroup
\renewcommand{\arraystretch}{1.25}%
\begin{table*}[t]
\caption{Test RMSE and standard errors for G-GLN versus three previously published methods on a standard suite of UCI regression benchmarks with $N$ instances of $d$ features each. Models are trained for 40 epochs and results summarized for 20 random seeds (5 for Protein).}
\centering
\begin{tabular}{lrrr@{$\pm$}l@{}r@{$\pm$}l@{}r@{$\pm$}l@{}r@{$\pm$}l@{}r@{$\pm$}l@{}rrr}
\textbf{Dataset} & $N$ & $d$
    &\multicolumn{2}{c}{\bf{ G-GLN }}&\multicolumn{2}{c}{VI~\cite{graves2011practical}}&\multicolumn{2}{c}{PBP~\cite{hernandez2015probabilistic}}&\multicolumn{2}{c}{DO~\cite{gal2015dropout}} \tabularnewline
\toprule
Boston Housing & 506 & 13
   &\bf{2.84}&0.03\;\; & 4.32&0.29\;\; & 3.01&0.18\;\; &2.97&0.19\tabularnewline
Concrete Compression Strength & 1030 & 8
   &5.84&0.03 & 7.13&0.12 & 5.67&0.09 &\bf{5.23}&0.12\tabularnewline
Energy Effiency & 768 & 8
   &\bf{1.31}&0.01 & 2.65&0.08 & 1.80&0.05 &1.66&0.04\tabularnewline
Kin8nm & 8192 & 8
   &\bf{0.09}&0.00 & 0.10&0.00 & 0.10&0.00 &0.10&0.00\tabularnewline
Naval Propulsion & 11,934 & 16
   &\bf{0.00}&0.00 & 0.01&0.00 & 0.01&0.00 &0.01&0.00\tabularnewline
Combined Cycle Power Plant & 9568 & 4
   &\bf{3.90}&0.01 & 4.33&0.04 & 4.12&0.03 &4.02&0.04\tabularnewline
Protein Structure & 45,730 & 9 
   &\bf{3.77}&0.01 & 4.84&0.03 & 4.73&0.01 &4.36&0.01\tabularnewline
Wine Quality Red & 1599 & 11
   &\bf{0.57}&0.00 & 0.65&0.01 & 0.64&0.01 &0.62&0.01\tabularnewline
Yacht Hydrodynamics & 308 & 6
   &3.76&0.04\;\; & 6.89&0.67 & \bf{1.01}&0.05 &1.11&0.09\tabularnewline
\bottomrule
\end{tabular}
\label{tab:uci}
\end{table*}
\endgroup

We applied G-GLNs to univariate regression, multivariate regression, contextual bandits with real valued rewards, denoising and image infilling. 
Model, experimental and implementation details common across our test domains are discussed below. 

\paragraph{Training Setup.}
Weights for all neurons in layer $i$ are initialized to $1/K_{i-1}$ where $K_{i-1}$ is the number of neurons in the previous layer.
Note that due to the convexity of the loss, the choice of initial weights plays a less prominent role in terms of overall performance compared with typical deep learning applications.
The only source of non-determinism in the model is the choice of context function; to address this, all of our results are reported by averaging over multiple random seeds.
For regression experiments, multiple epochs of training are used. Training data is randomly shuffled at the beginning of each epoch, and each example is seen exactly once within an epoch.

\paragraph{Bias and Base Predictions.}
Constant bias inputs for each neuron were set to span the target range, as described in Section~\ref{sec:neuron}. Given $d$-dimensional input of the form $(x_1, x_2, \dots, x_d)$, we adopted the convention of adding $d$ Gaussian PDF \emph{base predictions} to layer 0. The mean and variance of the $j$th expert was calculated online either by setting each expert to be centered at a single input feature $x_j$ and with a fixed width $\sigma$ such as $1.0$, or from an analytic formula that applies Bayesian Linear Regression (BLR) to learn a mapping of $x_j$ to an approximation of the target distribution.

\paragraph{Output Aggregation and Weight Projection.}
As each neuron in a G-GLN models the target distribution, any choice of neuron to be the output provides an estimate of the target density; we either take the output of the top-most neuron or use the switching aggregation method introduced in \cite{veness2017online} for B-GLNs which uses Bayesian tracking \cite{VenessNHB12} to estimate the best performing neuron on recent data. See Appendix~\ref{sec:switchingdetails} for details of switching aggregation.

We explored multiple methods for implementing weight projection efficiently, and obtained the best performance in our regression benchmarks by an approximate solution which used the log-barrier method~\cite{Boyd04}. This method essentially amounts to adding an additional regularization term to the loss, which has negligible affect on the cost of inference; see Appendix~\ref{app:projection} for implementation details.

\subsection{UCI Regression}
\label{subsec:uci}
We begin by evaluating the performance of a G-GLN to solve a benchmark suite of univariate UCI regression tasks. We adopt the same datasets and training setup described in~\cite{gal2015dropout}, and compare G-GLN performance to the previously published results for 3 MLP-based probabilistic methods: variational inference (VI)~\cite{graves2011practical}, probabilistic backpropagation (PBP)~\cite{hernandez2015probabilistic} and the interpretation of dropout (DO) as Bayesian approximation as described in~\cite{gal2015dropout}.  Our results are presented in Table~\ref{tab:uci}. It is evident that G-GLN achieves competitive performance, outperforming PBP, BP and DO on 7 out of 9 regression tasks. See Appendix~\ref{sec:experimental_methods_uci} for full details.

\begin{table}[t!]
    \caption{(\textbf{Left}) Test MSE for G-GLN versus previously published methods on the SARCOS inverse dynamics dataset~\cite{vijayakumar2000locally}. G-GLNs are trained for 2000 epochs using the same test procedure as~\cite{arik2019tabnet}. (\textbf{Right}) Performance of a G-GLN based GLCB algorithm for the continuous contextual bandits tasks and competitors described in~\cite{sezener2020online,riquelme2018deep}. Ranks are computed by running each algorithm on 500 randomly sampled environments. Raw scores are provided in Table~\ref{tab:bandits} of the Appendix.}
    \label{tab:sarcos}
    \begin{tabular}{ll}
    \bf{Algorithm}         & MSE  \\ \toprule
    \bf{G-GLN}                    &\textbf{0.10}       \\ 
    \midrule
    Random forest            & 2.39     \\
    MLP                      & 2.13                           \\
    Stochastic decision tree & 2.11                           \\
    Gradient boosted tree    & 1.44                           \\
    TabNet-S                 & 1.25        
       \\
    Adaptive neural tree     & 1.23                           \\
    TabNet-M                 & 0.28        
       \\
    TabNet-L                 & 0.14
       \\
    &\\
    \bottomrule
    \end{tabular} 
    \begin{tabular}{lllll}
    \textbf{Algorithm} & financial & jester & wheel & mean rank \\
    \toprule

    \bf{G-GLN}         &    3 &       \bf{1} &      2 & \bf{2}
    \\
    \midrule
    BBAlphaDiv   &     10 &       9 &     10 &9.67\\
    constSGD     &      9 &       8 &      6 &7.67\\
    ParamNoise   &      7 &      10 &      4 & 7 \\
    BBB          &      8 &       5 &      6  & 6.33\\
    NeuralGreedy &      5 &       4 &      9   & 6\\
    BootRMS      &      4 &       2 &      8  & 4.67\\
    Dropout      &      6 &       3 &      5  & 4.67\\
    NeuralLinear &      2 &       7 &      3  & 4\\
    LinFullPost  &      \bf{1} &       6 &      \bf{1} & 2.67\\
    \bottomrule
    \end{tabular}
\end{table}

\subsection{Inverse Dynamics}
\label{subsec:sarcos}
Next we demonstrate G-GLNs on regression tasks where both the inputs and targets are multi-dimensional. We consider the SARCOS dataset for a 7 degree-of-freedom robotic arm~\cite{vijayakumar2000locally}: using a 21-dimensional feature vector (7 joint positions, velocities and accelerations) to predict the 7 joint torques. We compare our performance to the state-of-the-art TabNet model~\cite{arik2019tabnet} and the same suite of standard regression algorithms considered by the TabNet authors. See Appendix~\ref{sec:SARCOS_details} for details.

Table~\ref{tab:sarcos} (left) shows that G-GLN outperforms all of the baselines, including the largest TabNet, which is a complex system of neural networks optimized for tabular data, exploiting residual transformer blocks for sequential attention. It is likely that a similar system could exploit G-GLNs as components for improved performance, but doing so is beyond the scope of this paper.

\subsection{Online Contextual Bandits}
\label{subsec:bandits}
The authors of~\cite{sezener2020online} proposed an algorithm, Gated Linear Context Bandits (GLCB), by which B-GLNs could be applied to solve contextual bandits tasks with binary rewards. GLCB provides a UCB-like~\cite{auer2002ucb} rule that exploits GLN half-space activation as a ``pseudo-count” that is shown to be effective for exploration (full details in Appendix~\ref{ss:bandits}). Our G-GLN provides a natural solution for extending GLCB to continuous rewards. Table~\ref{tab:sarcos} (right) compares the results of a G-GLN based GLCB algorithm (see Appendix~\ref{sec:bandit_details} for details) to three bandits tasks derived from UCI regression datasets, a standard benchmark in previous literature. G-GLN obtains the best mean rank across these tasks compared to 9 popular Bayesian deep learning methods~\cite{riquelme2018deep}. Similar to~\cite{sezener2020online}, our results are obtained in an online regime -- each data point is considered once without storage, whereas all other methods were able to i.i.d. resample from prior experience to learn an effective representation.

\subsection{Application to Denoising Density Estimation}
\label{subsec:denoising}
\begin{figure}[t]
\includegraphics[width=\linewidth]{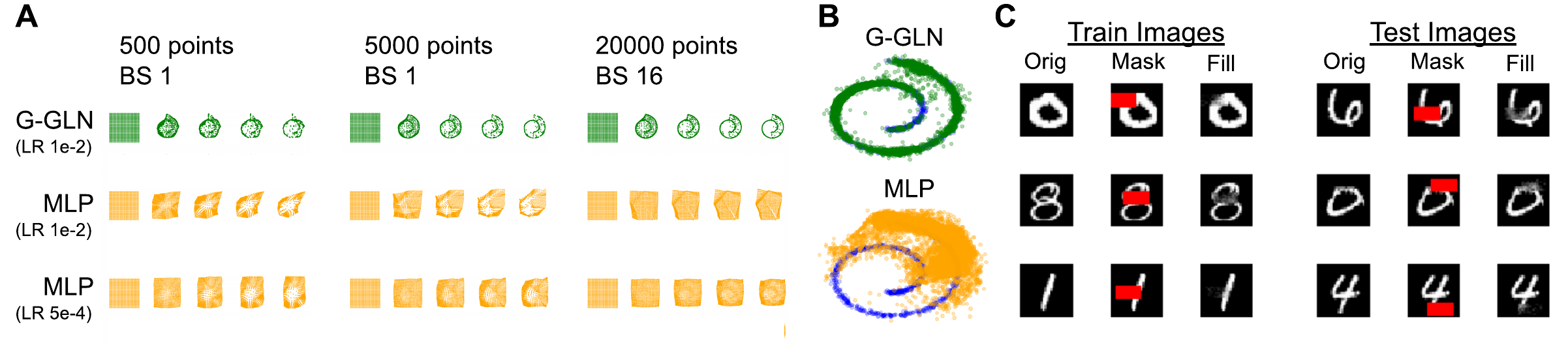}
\caption{Denoising multi-dimensional data with G-GLNs. (\textbf{A}) G-GLNs (top row) and MLPs (bottom two rows) are trained on 1-step denoising of a Swiss Roll density under additive Gaussian noise (BS = batch size, LR = learning rate). Starting with a grid, the original Swiss Roll data manifold is reconstructed with multi-step denoising. Larger version in Appendix~\ref{sec:bigdenoisingfig}. (\textbf{B}) Sampling via HMC using the gradient field inferred by denoising. Shown are samples from G-GLN inferred gradient (green), MLP inferred gradient (orange) and original data manifold (blue). (\textbf{C}) Infilling of MNIST train images (left) or unseen test images (right) is shown for binary occlusion masks, after training a G-GLN for only one epoch over the dataset with batch size 1 to remove additive Gaussian noise from each train image. Orig: original image. Mask: masked: Fill: filled. More examples in Appendix~\ref{sec:moremnistsupp}}
\label{fig:denoising}
\end{figure}

One application of high-dimensional regression is to the problem of density estimation via denoising~\cite{vincent2011connection, bigdeli2020learning, sohl2015deep, saremi2019neural}, which gives the ability to sample any conditional distribution from a learnt gradient of the log-joint data distribution. 
We use G-GLNs to approximate this score function, $\nabla_x \log p(x)$ \cite{hyvarinen2005estimation}, by using a G-GLN multivariate regression model as a denoising autoencoder \cite{vincent2011connection, bigdeli2020learning, sohl2015deep, saremi2019neural}. We train the GLN by adding isotropic Gaussian noise with covariance $\lambda \cal{I}$ ($0 < \lambda \ll 1$) to each data point and regressing to the un-noised point. At convergence, the vector $(x - \mu_{L,1}(x)) / \lambda$ approximates the score function \cite{alain2014regularized}, which we can feed into Hamiltonian Monte Carlo (HMC)~\cite{neal2011mcmc} to approximately sample from the distribution implied by the score field. See Appendix~\ref{sec:denoisingderivation} for details.

From Figure~\ref{fig:denoising}(A) it is evident that G-GLNs can learn reasonable approximate gradient fields for 2D distributions from just a single online pass of 500-5000 samples. Starting from a grid, multi-step denoising can then by applied to reconstruct the original data manifold. MLPs trained with the same data required a larger batch size and many more samples to accurately approximate the data density. This is evident in Figure~\ref{fig:denoising}(B), which shows the result of HMC sampling~\cite{neal2011mcmc} using the G-GLN versus MLP estimated gradient fields. Figure~\ref{fig:denoising}(C) demonstrates that the same process can be extended to much higher-dimensional problems, e.g. MNIST density modelling: iterative G-GLN denoising can be leveraged to fill in occluded regions in MNIST train or unseen test images after a single online pass through the train set in which it is trained to remove small additive Gaussian noise patterns from each image. This suggests an exciting avenue for future work applying G-GLNs as data-efficient pattern completion memories.

\section{Conclusion}
We have introduced a new backpropagation-free deep learning algorithm for multivariate regression that leverages local convex optimization and data-dependent gating to model highly non-linear and heteroskedastic functions.
We demonstrate competitive or state-of-the-art performance on a comprehensive suite of established benchmarks.
The simplicity and data efficiency of the G-GLN approach, coupled with its strong performance in high-dimensional multivariate settings, makes us optimistic about future extensions to a broad range of applications.


\clearpage
\section*{Software}
All models implemented using JAX~\cite{jax2018github} and the DeepMind JAX Ecosystem~\cite{chex2020github,haiku2020github,optax2020github,rlax2020github}. Open source GLN implementations (including G-GLN) are available at:
\newline\vspace{2mm}\texttt{www.github.com/deepmind/deepmind-research}.

\section*{Broader Impact}

Regression models have long been ubiquitous in both industry and academia, and we are optimistic that our work can provide improvement to existing practice and results. Like any supervised learning technique, the output of this model is a function of its input data, so appropriate due diligence is required during all stages of data collection, training and deployment, e.g. with respect to issues of algorithmic fairness and bias, as well as safety and robustness.

\section*{Acknowledgments}
We thank Agnieszka Grabska-Barwinska, Chris Mattern, Jianan Wang, Pedro Ortega and Marcus Hutter for helpful discussions.

\section*{Funding Disclosure}
All authors are employees of DeepMind.

\bibliographystyle{unsrt}

\clearpage
\appendix

\section{Weighted Products of Gaussians}
\label{app:pog_tutorial}

A well-known result is that a product of Gaussian PDFs collapses to a scaled Gaussian PDF (e.g.~\cite{bromiley2018}).
In particular, if we define
\begin{equation}\label{eq:gaussin_magic}
\sigma^2_\textsc{PoG} := \left[ \sum_{i=1}^m  \frac{1}{\sigma^2_i} \right]^{-1}
\text{~~~~and~~~~}
\mu_\textsc{PoG} := \sigma^2_\textsc{PoG} \left[ \sum_{i=1}^m \frac{\mu_i}{\sigma^2_i} \right],
\end{equation}
and let $f_\textsc{PoG}(\cdot)$ denote the associated PDF of $\mathcal{N}(\mu_\textsc{PoG}, \sigma_\textsc{PoG})$, then we have that
$f_\textsc{PoG}(y) \propto \prod_{i=1}^m f_i(y) $.
In the case where $w = \vec{1}$, this implies that $\text{\sc PoG}_{\vec{1}}(y \,;\, \dots) = f_\textsc{PoG}(y)$
as the constant of proportionality (not a function of $y$) is cancelled out by the division by Z in Equation \ref{eq:pog}, and we are left with an integral of a PDF in the denominator.
Now consider a Gaussian PDF $f$ raised to a power $p \in \mathbb{R}_+$, i.e.
\begin{equation*}
f^p(y) = \left [ \frac{1}{\sigma \sqrt{2 \pi} } \exp \left \{- \frac{1}{2} \left(\frac{y - \mu}{\sigma} \right)^2 \right \} \right]^p
\propto
\exp \left \{- \frac{1}{2} \frac{\left(y - \mu \right)^2}{\sigma^2 \, p^{-1}}  \right \},
\end{equation*}
which corresponds to an unnormalized Gaussian PDF with mean $\mu$ and variance $\sigma^2 \, p^{-1}$.
Thus we can replace each $f_i(y)^{w_i}$ term in Equation \ref{eq:pog} with the PDF associated with $\mathcal{N}(\mu, \sigma^2 \, p^{-1})$. 
Combining the above techniques for products and powers allows us to exactly interpret the weighted product of experts as another Gaussian expert $\mathcal{N} \left(\mu_\textsc{PoG}(w), \sigma^2_\textsc{PoG}(w) \right)$
where
\begin{equation}
\sigma^2_\textsc{PoG}(w) := \left[ \sum_{i=1}^m  \frac{w_i}{\sigma^2_i} \right]^{-1}
\text{~~~~and~~~~}
\mu_\textsc{PoG}(w) := \sigma^2_\textsc{PoG}(w) \left[ \sum_{i=1}^m \frac{w_i \, \mu_i}{\sigma^2_i} \right].
\end{equation}

\section{Properties of the G-GLN Loss}
\label{app:properties_of_loss}

\paragraph{Gradient.}
First define $\omega_i := w_i/ \sigma_i^2$, $\omega = (\omega_1, \dots, \omega_m)$ and $\mu = (\mu_1, \dots, \mu_m)$, which due to the non-negativity  of $w_i$ implies $\norm{\omega}=\sum_i \omega_i$.
Hence $\sigma^2_\textsc{PoG}= \norm{\omega}^{-1}$ and $\mu_\textsc{PoG} = {\omega^T \mu} \,/\, \norm{\omega}$. 
Using this notation, we can reformulate Equation \ref{eq:loss} as 
\begin{equation}\label{eq:vector_loss}
\ell(y ; \omega) {=} - \log \norm{\omega}  + {\left( x - \omega^T \mu / \wsum \right)^2}{\wsum} \: .
\end{equation}
The first partial derivative can be obtained by direct calculation, and is
\begin{equation*}
\frac{\partial \ell(y ; \cdot)}{\partial \omega_i} = - \norm{\omega}^{-1}  + \left( y - \omega^T \mu / \wsum \right) \left( y - 2 \mu_i + \omega^T \mu / \wsum \right) .
\end{equation*}
Hence, using the above and $\frac{\partial \ell(y ; w)}{\partial  w_i} = \frac{\partial \ell(y ; \omega)}{\partial \omega_i} \frac{\partial \omega_i}{\partial w_i} = \frac{\partial \ell(y ; \omega)}{\partial \omega_i} \frac{1}{\sigma^2_i}$, we have
\begin{align*}
\nabla_{w} \, \ell(y;w) &=  \diag \left(\tfrac{1}{\sigma^2}\right) \left[ (y - \mu_\textsc{PoG} ) \left( \mathds{1}_{m,1} (y + \mu_\textsc{PoG} ) - 2 \mu \right) - \mathds{1}_{m,1} \, \sigma^2_\textsc{PoG} \right].
\end{align*}

\paragraph{Convexity.} 
Here we prove that $\ell(y ; w)$ is a convex function of $w$ by showing that the Hessian of Equation \ref{eq:vector_loss} is positive semi-definite (PSD).
Let $g(\omega) := \omega^T \mu \,/\, \wsum$ and $
g'(\omega) :=\frac{\partial g}{\partial \omega_j}= \mu_j \norm \omega^{-1} - \omega^T \mu \norm \omega^{-2}$, which allows us to compute the second partial derivative as
\begin{align*}
\frac{\partial^2 \ell(y ; \cdot)}{\partial \omega_i \partial \omega_j} &= \norm{\omega}^{-2} - g'(\omega)y + 2g'(\omega)\mu_i + g'(\omega)y - 2g(\omega)g'(\omega) \\
&= \norm{\omega}^{-2} + 2 g'(\omega) (\mu_i - g(\omega))\\
&= \norm{\omega}^{-2} + 2 (\mu_j \wsum^{-1} - \omega^T \mu \wsum^{-2})(\mu_i - \omega^T \mu / \wsum)\\
&= \norm{\omega}^{-2} + 2 \wsum^{-1} (\mu_j - \omega^T \mu /\wsum)(\mu_i - \omega^T \mu / \wsum)\: .
\end{align*}
Thus the Hessian of Equation \ref{eq:vector_loss} is
\begin{equation}\label{eq:hessian_loss}
    \nabla^2 \ell(y ; \omega) = \norm \omega^{-2} \mathds{1}_{m, m} + 2 \norm \omega^{-1}(\mu - \omega^T \mu/\norm \omega \mathds{1}_{m,1} )(\mu - \omega^T \mu/\norm \omega \mathds{1}_{m,1})^T \: ,
\end{equation}
where $\mathds{1}_{m, n}$ denotes the $m \times n$ matrix whose entries are all 1.
As $\mathds{1}_{m, m}$ is PSD and $\norm \omega^{-2} > 0$, the first additive term is PSD.
The second term is also PSD, since $2 \norm \omega^{-1} > 0$ and the outer product $(\mu - \omega^T \mu/\norm \omega \mathds{1}_{m,1} )(\mu - \omega^T \mu/\norm \omega \mathds{1}_{m,1})^T$ is PSD by letting $a=(\mu - \omega^T \mu/\norm \omega \mathds{1}_{m,1} )$ and observing that $u^\top a a^\top u = (u^\top a) (u^\top a)^\top = (u \cdot a)^2 \geq 0$ for all $u \in \mathbb{R}^m$.
Hence since the Hessian is the sum of two PSD matrices, it is PSD which implies that $\ell(y ; \omega)$ and therefore $\ell(y ; w)$ is a convex function of $w$.

\section{Learning the Base Model}
\label{sec:blr}
Every neuron in a G-GLN takes one-or-more Gaussian PDFs as input and produces a Gaussian PDF as output. This raises the question of what input to provide to neurons in the first layer, i.e. the \emph{base prediction}. We consider three solutions: (1) None. The input sufficient statistics to each neuron are already concatenated with so-called ``bias” Gaussians to ensure that the target mean falls within the convex hull defined by the input means (described in Section~\ref{sec:neuron}). (2) A Gaussian PDF for each component $x_i$ of the input vector, with $\mu=x_i$ and $\sigma =$ constant. It is perhaps surprising that the neuron inputs are not required to be a function of the $x_i$s, but this is permissible because $x_i$ is z-score normalized and broadcast to every neuron as side information $z_i$.

\begin{figure}[t]
\includegraphics[width=\linewidth]{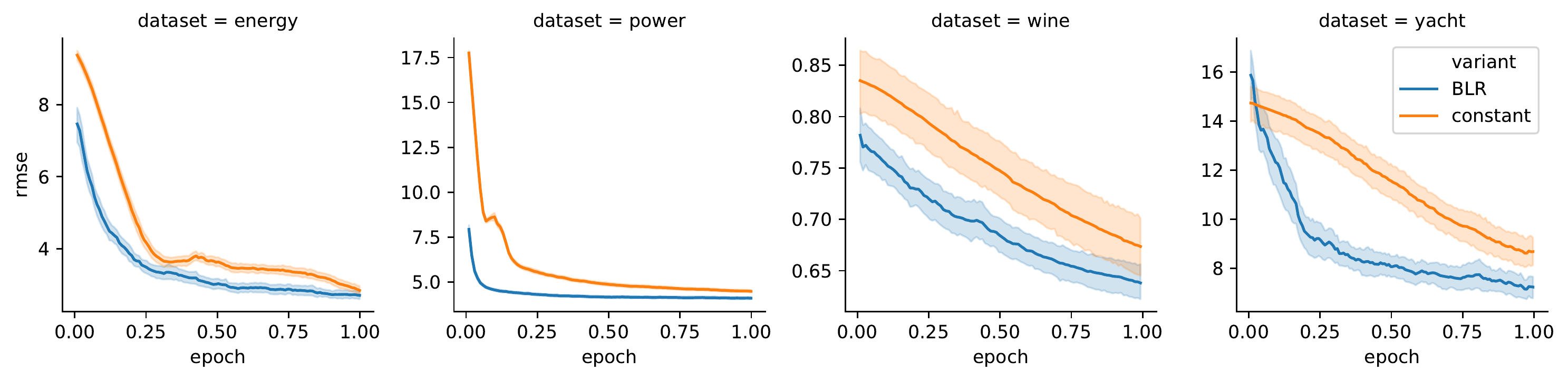}
\caption{Effect of using Bayesian linear regression (BLR) versus a constant base model $\mathcal{N}(0, 1)$ on predictive RMSE for four UCI regression tasks. Results are shown for the first epoch of training.}
\label{fig:blr}
\end{figure}

We present a third option (3) whereby the base prediction is provided by a probabilistic \emph{base model} trained to directly predict the target using only a single feature dimensions. The formulation of this Bayesian Linear-Gaussian Regression (BLR) model is described below. Empirically we find that it leads to improved data efficiency in the first epoch of training (see examples in Figure~\ref{fig:blr}) with only an additional $\mathcal{O}(1)$ time and space cost per feature dimension.

Consider a dataset $\mathcal{D}=\{x_i, y_i\}_{i=1}^N$ of zero-centered univariate features $x_i \in \mathbb{R}$  and corresponding targets $y_i \in \mathbb{R}$. We assume a Normal-linear relationship between a feature $x_i$ and target $y_i$,
\[
   y_i \sim \mathcal{N}(\theta x_i + \beta, \tau^{-1})
\]
where $\theta$ and $\beta$ are some coefficients, and $\tau$ is the precision (inverse variance). 
We assume $\tau$ is known, but it can also be optimized via (type II) maximum likelihood estimation.
We also assume an isotropic Normal prior over $\theta$ and $\beta$, i.e. $\theta \sim \mathcal{N}(0, \tau_0^{-1})$ and $b \sim \mathcal{N}(0, \tau_0^{-1})$, where $\tau_0$ is the prior precision.

By adapting widely known equations (e.g. Equations 3.53-3.54 in \cite{bishop2006prml}) we can obtain the posterior for $\theta$ as
\begin{align*}
    p(\theta \mid \mathcal{D}) &= \mathcal{N}(\theta \mid \mu_\theta, \tau_\theta^{-1}) \\
    \mu_\theta &= \tau \tau_\theta^{-1} \sum_{x_i, y_i \in \mathcal{D}} x_i y_i \\
    \tau_\theta &= \tau_0 + \tau \sum_{x_i \in \mathcal{D}} x_i^2 \: .
\end{align*}

Similarly, we obtain the posterior for $\beta$ as
\begin{align*}
    p(\beta \mid \mathcal{D}) &= \mathcal{N}(\beta \mid \mu_\beta, \tau_\beta^{-1}) \\
    \mu_\beta &= \tau \tau_\beta^{-1} \sum_{y_i \in \mathcal{D}} y_i \\
    \tau_\beta &= \tau_0 + \tau N \: .
\end{align*}

Putting these two together, we can obtain the posterior predictive distribution,
\[
    p(y \mid x, \mathcal{D}) = \mathcal{N}(y \mid \mu_\theta x + \mu_\beta,  x^2 \tau_\theta^{-1} +  \tau_\beta^{-1} + \tau^{-1}) \: .
\]

It is apparent that updates and inference can be performed incrementally in constant time and space by storing and updating the sufficient statistics $\sum_i x_i y_i$, $\sum_i x_i^2$, $\sum_i y_i$, $\sum_i 1$.

We can use this BLR formulation to convert the input features into probability densities. Specifically, for each feature, we independently maintain posterior/sufficient statistics and use the posterior predictive distributions as inputs to the base layer of the G-GLN.

\section{Switching Aggregation}
\label{sec:switchingdetails}
Because every neuron in a G-GLN directly models the target distribution, there is no one natural definition of the network output. One convention is simply to have a final layer consisting of a single neuron, and take the output of that neuron as the network output. An alternative method of \emph{switching aggregation} was used in~\cite{veness2017online,veness2012context}, whereby an incremental online update rule was used to weight the contributions of individual neurons in the network to an overall estimate of the target density.

We extend the switching aggregation procedure from the Bernoulli to Gaussian case by replacing a Bernoulli target probability value with a Gaussian probability density value evaluated at the target. The switching algorithm of~\cite{veness2012context} was originally presented in terms of log-marginal probabilities, which can cause numerical difficulties at implementation time. Instead we use an equivalent formulation derived from \cite{veness2017online} that incrementally maintains a weight vector that is used to compute a convex combination of model predictions, i.e. the densities given by each neuron in the network, at each time step. 

Using notation similar to~\cite{veness2017online}, let $m \geq 2$ denote the number of neurons, and $w_t^i \in [0,1]$ denote the weight associated with model $i$ at times $t \geq 1$.
The density output by the $i$th neuron at time $t$, evaluated on target $y_t$, will be denoted by $\rho_i(y_t \; | \; y_{<t})$.
At each time step $t$, switching aggregation outputs the density
\begin{equation*}
\pi(y_t \; | \; y_{<t}) := \sum_{i=1}^m w^i_t \; \rho(y_t \; | \; y_{<t}),
\end{equation*}
with the weights defined, for all $1 \leq i \leq m$, by $w^i_1 := 1 / m$ and
\begin{equation*}
w^i_{t+1} = \frac{\alpha_{t+1}}{m-1} +
\left( (1-\alpha_{t+1}) - \frac{\alpha_{t+1}}{m-1} \right) \frac{w^i_t \; \rho_i(y_t | y_{<t})}{\pi(y_t \; | \; y_{<t})} ,
\end{equation*}
with $\alpha_t := 1/t$.
This can be implemented in linear time with respect to the number of neurons.
Notice that mathematically the weights satisfy the invariant $\sum_{i=1}^m w_t^i = 1$ for all times $t \geq 1$, which should be explicitly enforced after each update to avoid numerical issues in any practical implementation.

\section{Weight Projection}
\label{app:projection}
Weight projection after an update (Line 11 in Algorithm~\ref{alg:ggln}) enforces three sets of constraints: each weight to be in $[0, b]$, mixed means $\mu_\textsc{PoG}$ to be in $[\mu_{\min}, \mu_{\max}]$, and mixed variances $\sigma^2_\textsc{PoG}$ to be in  $[\sigma^2_{\min}, \sigma^2_{\max}]$. These constraints ensure that the online convex optimization is well-behaved by forming a convex feasible set and also preventing numerical issues that arise from rounding likelihoods $\mathcal{N}(x; \mu_\textsc{PoG}, \sigma_\textsc{PoG})$ to $0$. We outline two ways in which these constraints can be implemented below.

The constraints can be represented in terms of linear inequalities $A w \le u$, where $w = W_{ij c_{ij}(z)}$ is the weight vector of neuron $\langle i, j \rangle$ given side info $z$. Assume $w$ violates some of the constraints, therefore we would like to project $w$ onto our feasible set $\{w': Aw' \le u\}$. Let $A'$ and $u'$ be the matrix/vector composed of rows/elements of $A$ and $u$ respectively that violate our original inequality, thus $A' w > u'$. Then we can write down the projection problem as $\arg \min_{w'} ||w' - w||^2$ s.t. $A' w' = u'$, the solution of which is $w - A^\dagger (A'w - u')$ where $A^\dagger = A'^T (A' A'^T)^{-1}$ is the pseudo-inverse of $A'$. This pseudo-inverse can be computed efficiently, because all but (at most) two rows of $A'$ are ``one-hot''.

The exact projection approach relies on dynamically shaped $A'$ and $u'$, support for which is limited in contemporary differentiable programming libraries such as Tensorflow~\cite{abadi2016tensorflow} and JAX~\cite{jax2018github}. Therefore, we take an alternative approach and enforce the inequalities via using logarthmic barrier functions (log-barriers) that augment the original loss function by penalizing the weights that are close to the constraints. Let $A_k$ and $u_k$ be the $k$th row and element of $A$ and $u$ respectively. For the constraint $A_k^T w \le u_k$, we can define a barrier function 
\[\phi_k(w) =
\begin{cases} 
      -\log(u_k - A_k^T w) & A_k^T w < u_k \\
      +\infty & \textrm{otherwise}
   \end{cases} \: .
\]
Note that we are now dealing with strict inequalities rather than $\le$ for convenience.
We can then augment the loss function $\ell(y; w)$ from Equation~\ref{eq:loss}, incorporating the barriers,
\begin{equation}
    \ell_{\textrm{combined}}(y; w) = \ell(y; w) + \xi \Phi(w)
\end{equation}
where $\Phi(w) = \sum_k \phi_k(w)$ and $\xi > 0$ is the barrier constant. 
Note that $\ell_{\textrm{combined}}(y; w)$ is convex in $w$ as each $\phi_k(w)$ is convex.

The weight updates can be carried out via $w \leftarrow w - \eta \nabla  \ell_{\textrm{combined}}(y; w)$. For sufficiently small $\eta$ and sufficiently large $\xi$, we will not need the projection step in Line 11 of Algorithm~\ref{alg:ggln}, as the constraints are incorporated into the loss function. However, in practice, we need backstops in case weights pass through the barriers due to large gradient steps. We implement the backstops by first hard-clipping each weight to be in $[0, b]$ then by enforcing $\sigma_{\textrm{min}}^{-2} > \sigma_{\textsc{PoG}}^{-2} = w^T \sigma_i^{-2} > \sigma_{\textrm{max}}^{-2}$, which corresponds to performing a single linear projection if the inequality is violated.

\section{Denoising Density Estimation}
\label{sec:denoisingderivation}
With $\hat{p}$ denoting a Gaussian likelihood function (as parameterized by a G-GLN) and $p^d(x)$ an unknown data-generating distribution, suppose we add isotropic Gaussian noise of variance $\lambda$ to sampled data points and then denoise them back to the original samples. The expected loss is
\begin{align}
& \expect{x \sim p^d(x)}{\expect{\xi \sim \mathcal{N}(0, \lambda)}{\ln \hat{p}(x \mid z, x + \xi)}} \nonumber \\
& = \expect{x \sim p^d(x)}{\expect{\xi \sim \mathcal{N}(0, \lambda)}{\ln \frac{\exp(-\|x - \mu(x + \xi) \|^2 / (2 \sigma^2))}{(2 \pi \sigma^2)^{(d/2)}}}} \nonumber \\
& = \expect{x \sim p^d(x)}{\expect{\xi \sim \mathcal{N}(0, \lambda)}{-\|x - \mu(x + \xi) \|^2 / (2 \sigma^2(x + \xi)) - (d/2) \ln (2 \pi \sigma^2(x + \xi))}}. \nonumber
\end{align}
Taking the variational derivative of this expected loss with respect to our G-GLN demonstrates the relationship between the value of the optimal output $\mu(x)$ and the gradient of the log data density:
\begin{align}
0 & = \expect{\xi}{p^d(x - \xi) (x - \xi - \mu(x))} \nonumber \\
& = \expect{\xi}{(p^d(x)  - \nabla_{x} p^d(x) \cdot \xi + \mathcal{O}(\|\xi\|^2)) (x - \xi - \mu(x))} \nonumber \\
\implies \mu(x) & = \frac{p^d(x) x  + \nabla_{x} p^d(x) \lambda }{p^d(x)} \nonumber \\
& = x + \lambda \nabla_{x} \ln p^d(x),
\end{align}
in the limit $||\xi||_2 \to 0$.
Therefore, we can approximate the gradient field as $(\mu(x) - x)/\lambda$, which we use in the main text. Hamiltonian Monte Carlo sampling then takes as input this gradient estimate for $\nabla_x \ln p^d(x)$. Denoising iteratively applies the G-GLN, trained on denoising, to an arbitrary starting point $x \to \mu(x) \to \mu(\mu(x))$, and so on. 

\section{Additional Results}

\subsection{Contextual Bandits}
\label{ss:bandits}
\begingroup
\renewcommand{\arraystretch}{1.15}%
\begin{table*}[t!]
\caption{Performance of the GLN-based GLCB algorithms for the contextual bandits tasks and competitors described in~\cite{sezener2020online,riquelme2018deep}. G-GLCB uses a single G-GLN instead of a CTree of 7 equivalent-sized B-GLNs (italics), the method described in~\cite{sezener2020online}, to model continuous-valued results. Results are mean and standard error of cumulative rewards over $500$ random environment seeds.}
\begin{center}
\begin{tabular}{lccccccc}
\multicolumn{1}{c}{} & \multicolumn{4}{c}{Binary targets} & \multicolumn{3}{c}{Continuous targets}\\
\cmidrule(lr){2-5} \cmidrule(lr){6-8}
\bf{Algorithm} &      adult &       census &    covertype &      statlog &    financial &      jester &        wheel \\
\toprule
G-GLN       & - &   - &  - &  - &   3018$\pm$3 & \textbf{3301$\pm$4} &  4386$\pm$11 \\
B-GLN       & \textbf{678$\pm$5} &   \textbf{2718$\pm$3} &  2715$\pm$12 &  \textbf{4863$\pm$1} &   \emph{3038$\pm$3} & \emph{3298$\pm$3} &  \emph{4432$\pm$11} \\
\midrule
BBAlphaDiv   &   18$\pm$2 &   932$\pm$12 &   1838$\pm$9 &  2731$\pm$15 &   1860$\pm$1 &  3112$\pm$4 &  1776$\pm$11 \\
BBB          &  399$\pm$8 &  2258$\pm$12 &  2983$\pm$11 &  4576$\pm$10 &  2172$\pm$18 &  3199$\pm$4 &  2265$\pm$44 \\
BootRMS      &  676$\pm$3 &   2693$\pm$3 &  \textbf{3002$\pm$7} &  4583$\pm$11 &   2898$\pm$4 &  3269$\pm$4 &  1933$\pm$44 \\
Dropout      &  652$\pm$5 &   2644$\pm$8 &   2899$\pm$7 &  4403$\pm$15 &   2769$\pm$4 &  3268$\pm$4 &  2383$\pm$48 \\
LinFullPost  &  463$\pm$2 &   1898$\pm$2 &   2821$\pm$6 &   4457$\pm$2 &  \textbf{3122$\pm$1} &  3193$\pm$4 &  \textbf{4491$\pm$15} \\
NeuralGreedy &  598$\pm$5 &  2604$\pm$14 &   2923$\pm$8 &  4392$\pm$17 &   2857$\pm$5 &  3266$\pm$8 &  1863$\pm$44 \\
NeuralLinear &  391$\pm$2 &   2418$\pm$2 &   2791$\pm$6 &   4762$\pm$2 &   3059$\pm$2 &  3169$\pm$4 &  4285$\pm$18 \\
ParamNoise   &  273$\pm$3 &   2284$\pm$5 &   2493$\pm$5 &  4098$\pm$10 &   2224$\pm$2 &  3084$\pm$4 &  3443$\pm$20 \\
constSGD     &  107$\pm$3 &  1399$\pm$22 &   1991$\pm$9 &  3896$\pm$18 &   1862$\pm$1 &  3136$\pm$4 &  2265$\pm$31 \\
\bottomrule
\end{tabular}
\end{center}
\label{tab:bandits}
\end{table*}
\endgroup

In~\cite{sezener2020online} the authors present a B-GLN based algorithm, GLCB, that achieves state-of-the-art results across a suite of contextual bandits tasks with both binary and real-valued rewards. The former uses the B-GLN formulation directly. For the latter, the authors present an algorithm called CTree for tree-based discretization, i.e. using $b-1$ B-GLNS arranged within a binary tree structure to model the target distribution over $b$ bins. In both cases, GLCB leveraged properties of GLN half-space gating to derive a UCB-like \cite{auer2002ucb} rule based on ``pseudo-counts" (inspired by~\cite{bellemare2016unifying}) to help guide exploration. At each timestep $t$, the GLCB policy \cite{sezener2020online} greedily maximizes a linear combination of the expected action reward as predicted by a GLN and an exploration bonus $\sqrt{\log t / \hat{N}(s_t, a)}$ where $\hat{N}(s_t, a)$ is the pseudocount term capturing how similar the current context-action pair $\langle s_t, a \rangle$ is to the previously seen data. This term is computed at no additional cost by utilizing gating functions of GLN neurons.

Table~\ref{tab:bandits} expands on the results in Section \ref{subsec:bandits} to demonstrate the performance of GLNs for both binary and continuous-valued rewards. It is evident that GLNs achieve state-of-the-art performance in both regimes. Moreover, using the natural G-GLN formulation described in this paper is able to match the previous performance of a CTree of B-GLNs with just a single equivalent-sized network (an order-of-magnitude reduction in memory and computation cost).

\subsection{2D Denoising}
\label{sec:bigdenoisingfig}
Figure 4 shows 24 steps of denoising starting from a grid for the Swiss Roll gradient fields. At larger batch sizes and lower learning rates, and with more denoising steps (lower right panel), the MLP control begins to approximate the Swiss Roll data manifold.

\begin{figure}
\label{fig:big_denoising_blowup}
\includegraphics[width=\linewidth]{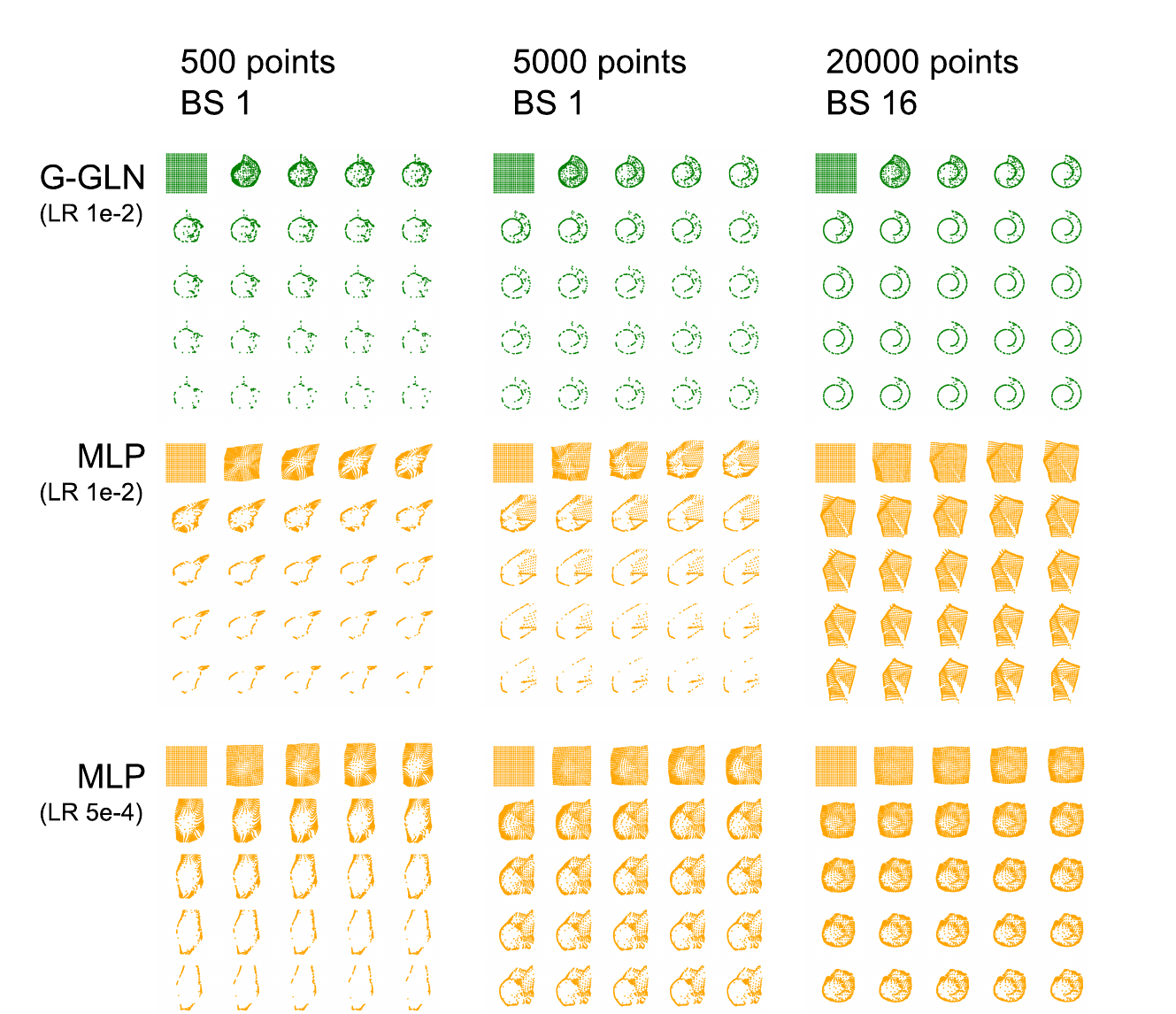}
\caption{G-GLNs (top set of rows) and MLPs (bottom two sets of rows) are trained on 1-step denoising of added Gaussian noise using data points sampled from a Swiss Roll. Subsequently, iterative multi-step denoising starting from a grid reconstructs an approximation of the original Swiss Roll data manifold. BS denotes batch size, LR denotes learning rate. The initial grid followed by 24 steps of denosing are shown left to right and top to bottom.}
\end{figure}

\subsection{MNIST Infilling}
\label{sec:moremnistsupp}
Figure 5 shows the result of 3000 steps of denoising of MNIST train and test digits, after training for 1 epoch at batch size 1. This shows that the network, which has been trained on denoising small additive Gaussian noise perturbations to train set digits, is able to denoise unseen binary mask perturbations on unseen test set digits. This occurs over many iterative steps of denoising, much as the grid in Figure 4 is iteratively denoised to the Swiss Roll data manifold.

\begin{figure}
\label{fig:more_mnist_supp}
\center
\includegraphics[width=0.5\linewidth]{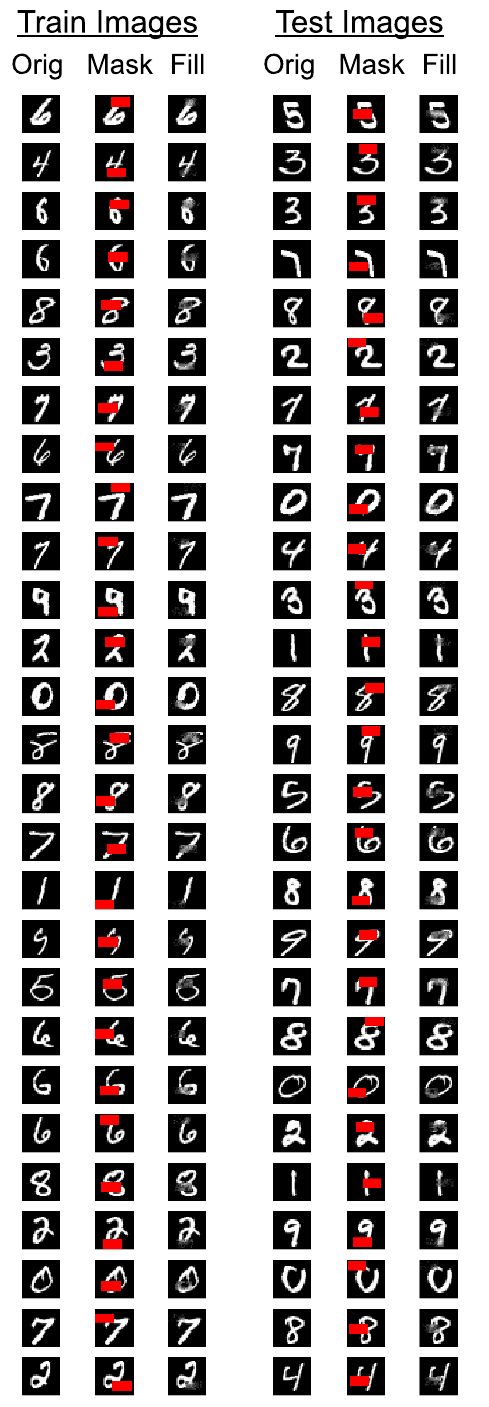}
\caption{Further MNIST infilling examples. G-GLN was trained for 1 epoch at batch size 1 by denoising a small additive Gaussian noise pattern from each train image. Subsequently, it can remove unseen binary occulsion masks either from train images (left) or unseen test images (right). Orig: original image. Mask: masked image: Fill: filled image. Examples were randomly chosen.}
\end{figure}

\section{Experimental Details}
\label{sec:experimental_methods}

\subsection{UCI regression details}
\label{sec:experimental_methods_uci}
Each G-GLN was trained with batch size 1 for 40 epochs of a randomly selected 90\% split of the dataset (except DO which was trained for 400). The predictive RMSE is evaluated for the remaining 10\%, with the mean and standard error reported across 20 different splits (5 for Protein Structure). Similarly to~\cite{hernandez2015probabilistic}, we normalize the input features and targets to have zero mean and unit variance during training. Target normalization is removed for evaluation. 

For each UCI dataset we train a G-GLN with 12 layers of 256 neurons. Context functions are sampled as described in Section~\ref{sec:ggln} with an additive bias of 0.05. The switching aggregation scheme was used to generate the output distribution. In~\cite{hernandez2015probabilistic} the authors specify that 30 configurations of learning rate, momentum and weight decay parameters are tuned for each task for VI, BP and PBP. We likewise search 12 configurations of learning rate $\in \{1e^{-3}, 3e^{-3}, 1e^{-2} \}$ and context dimension $\in \{4, 6, 8, 10\}$ for each task and present the best result. 

\subsection{SARCOS details}
\label{sec:SARCOS_details}
The G-GLN was trained for 2000 epochs using the SARCOS test and train splits defined in~\cite{vijayakumar2000locally}. Inputs were normalized to have zero mean and unit variance during training, with the target component-wise linearly rescaled to $[-1, 1]$. Fixed bias Gaussians were placed with means $\pm 7$ and variance 5 along each of the 7 output coordinate axes. The network base model uses Gaussians with standard deviation 1 centered on each component $x_i$ of the input vector.

The G-GLN was trained with 4 layers of 50 neurons, context dimension 14, and learning rate 0.01. Context functions are sampled as described in Section~\ref{sec:ggln} with an additive bias of 0.05. The switching aggregation scheme was used to generate the output distribution. We enforce weights to be in $[-10^5, 10^5]$ and mixed variances $\sigma^2_\textsc{PoG}$ to be in $[1, 10^9]$ by performing projections when needed.

\subsection{Contextual bandits details}
\label{sec:bandit_details}
We adopt the experimental configuration described in \cite{sezener2020online}, including inputs and target scaling and method of hyperparameter selection. Performance was evaluated across 500 seeds per dataset. The G-GLN was trained with shape $[1000, 100, 1]$ with context dimension 1 and a learning rate of 0.003. A single output layer with a single neuron was used to generate the output distribution. Context functions are sampled as described in Section~\ref{sec:ggln} with an additive bias of 0.05. For the GLCB algorithm a \textsc{UCB} exploration bonus of 1 was chosen with mean-based pseudo-count aggregation.

\subsection{Denoising details}
\label{sec:denoising_details}
The MLP control for Swiss Roll denoising was a ReLU network with hidden layer sizes 64 and 32 and output size 3 (2D $\mu$ and 1D $\sigma^2$). Both were trained with Gaussian log likelihood. The MLP was evaluated with learning rates of both $0.01$ or $0.0005$ for comparison. For Hamiltonian Monte Carlo (HMC) sampling, 15000 HMC steps were performed, with each step consisting of 150 sub-steps and $\epsilon=0.003$. No acceptance criterion was used. Particle mass was 1.

For the MNIST image denoising, the G-GLN was trained with 6 layers of batch size 50 with context dimension of 10 and a learning rate of 0.05. The network base model uses Gaussians with variance 0.3 centered on each component $x_i$ of the input vector. A single output layer with a single neuron was used to generate the output distribution. 

For MNIST denoising, context functions are sampled as described in Section~\ref{sec:ggln} with a normally distributed additive bias of scale 0.05, while for Swiss Roll denoising in 2D, the additive bias scale was 0.5 to ensure proper tiling of the low-dimensional input space with hyperplane regions. 

The G-GLN was trained in a single pass through all train points with batch size 1, with data represented as flat $28^2=784$ dimensional vectors. The model was trained to remove a single additive Gaussian noise pattern for each train image during training, and was then tested on MNIST in-filling using an independent test set of images occluded by unseen randomly positioned binary masks. To estimate a gradient direction for infilling, a single step of the trained denoising procedure was performed on each successive image, then a step of length 0.002 was taken interpolating between the image and the denoised prediction, after which pixels outside the masked region were projected back to their original values. This was repeated iteratively up to 3000 times. 

For both Swiss Roll and MNIST denoising, target data was component-wise linearly scaled to $[-1, 1]$. For MNIST, we first added Gaussian noise of standard deviation 75 to the first 10k train points to define an appropriate scaling range for the linear scaler. All weights were kept positive by clipping to a maximum of 1000. A minimum $\sigma^2$ was enforced by clipping during inference but not updating. Log-barriers were not used.
\end{document}